\documentclass[11pt]{article}

\usepackage[margin=1in]{geometry}
\usepackage[T1]{fontenc}
\usepackage[utf8]{inputenc}
\usepackage{lmodern}
\usepackage{microtype}
\usepackage{amsmath,amssymb,amsfonts}
\usepackage{booktabs}
\usepackage{enumitem}
\usepackage{xcolor}
\usepackage{hyperref}
\usepackage{url}
\usepackage{listings}
\usepackage{array}
\usepackage{longtable}
\usepackage{graphicx}

\hypersetup{
  colorlinks=true,
  linkcolor=blue!60!black,
  citecolor=blue!60!black,
  urlcolor=blue!60!black
}

\lstdefinestyle{ccl}{
  basicstyle=\ttfamily\small,
  breaklines=true,
  columns=fullflexible,
  frame=single,
  framerule=0.3pt,
  backgroundcolor=\color{gray!5}
}

\title{\textbf{Compress the Context, Keep the Commitments: A Formal Framework for Verifiable LLM Context Compression}}
\author{Natalia Trukhina \quad Vadim Vashkelis\\
Embedded Intelligence Lab (EMILAB)\\
\texttt{ntrukhina@emilab.org} \quad \texttt{vvashkelis@emilab.org}\\
\url{https://emilab.org}}
\date{}

\begin{document}
\maketitle

\begin{abstract}
LLM context is not just tokens; it is a set of commitments. Long-running conversations accumulate goals, constraints, decisions, preferences, tool results, retrieved evidence, artifacts, and safety boundaries that future responses must preserve. Existing context-management methods reduce length through truncation, retrieval, summarization, memory systems, or token-level prompt compression, but they rarely specify \textit{which semantic commitments must survive compression} or how their preservation should be measured. We propose \textit{Context Codec}, a commitment-level framework for compressing prompts and chat histories.

Context Codec represents dialogue state as typed, source-grounded semantic atoms with canonical identity, equivalence, conflict, confidence, risk, and evidence spans. It separates five concerns---extraction, normalization, representation, rendering, and verification---and introduces metrics for Critical Atom Recall, Weighted Atom Recall, Commitment Density, and round-trip recoverability. It also defines a taxonomy of semantic compression errors, a concrete normalization procedure, conservative fallback rules for low-confidence and safety-critical atoms, and Context Compression Language (CCL), an ASCII-first compact rendering of canonical JSON atoms. In a small diagnostic study, CCL-Core occupies a useful middle ground between structured prose and JSON: more explicit and auditable than prose, usually more compact than JSON, and less risky than heavily minified notation. The result is not a claim that shorthand solves compression, but a framework for making context compression verifiable: compress the conversation, keep the commitments.
\end{abstract}

\section{Introduction}

The context supplied to an LLM increasingly determines the behavior of the system around it. A modern assistant may rely on system instructions, user preferences, project history, source documents, retrieved passages, tool outputs, plans, intermediate artifacts, safety policies, and prior decisions. As context grows, systems must decide what to include, what to summarize, what to retrieve, and what to drop.

The naive solution is to increase the context window. However, long context alone does not solve context management. Prior work shows that LLMs may not use long contexts robustly: performance can depend on the position of relevant information, with evidence in the middle of long contexts being harder to use \cite{liu2024lost}. Long contexts also increase latency and cost, motivating prompt compression methods such as LLMLingua \cite{jiang2023llmlingua}, LongLLMLingua \cite{jiang2024longllmlingua}, Selective Context \cite{li2023selective}, and LLMLingua-2 \cite{pan2024llmlingua2}. Retrieval-augmented generation (RAG) combines parametric model knowledge with retrieved non-parametric evidence \cite{lewis2020rag}; memory systems such as MemGPT manage context using memory tiers \cite{packer2023memgpt}. These approaches are complementary to the present work.

This paper argues that context compression should be analyzed at the level of \emph{semantic commitments}. A semantic commitment is any proposition, constraint, decision, preference, state variable, or safety boundary that can change the correctness of a future response. Examples include: ``the output must be a single HTML file,'' ``no external libraries are allowed,'' ``deduplicate by customer id,'' ``the user rejected SVG and chose Canvas,'' or ``the open question is whether the notation should be ASCII-only.'' Losing one of these commitments may make a future answer wrong even if a summary remains fluent.

\paragraph{Running example: prompt compression.} Consider the following accumulated prompt state from a data-cleaning task:
\begin{lstlisting}[style=ccl]
Write a Python script that reads customers.csv and writes cleaned.csv.
Use only the standard library: no pandas and no third-party packages.
The input columns are customer_id, email, signup_date, country, and
revenue_usd. Drop rows with missing customer_id or invalid email. Normalize
country names to ISO-2 codes using a small in-file mapping. Parse dates in
YYYY-MM-DD, MM/DD/YYYY, and DD Mon YYYY formats and output ISO dates. Convert
revenue_usd to a decimal string with two places; invalid revenue becomes
0.00. Deduplicate by customer_id, keeping the row with the latest signup_date.
Print a final report with rows_read, rows_written, rows_dropped, and
duplicates_removed. The answer should be code only.
\end{lstlisting}
A commitment-level compressor does not try to preserve every sentence. It extracts the commitments that control future correctness and can render them compactly:
\begin{lstlisting}[style=ccl]
@CCL/1
TASK=code.python.csv_clean
IN=customers.csv
OUT={file:cleaned.csv,answer:codeonly}
C={stdlib_only:true,pandas:false,third_party:false}
COLS=[customer_id,email,signup_date,country,revenue_usd]
DROP={missing:customer_id,invalid:email}
NORM={country:iso2_map_in_file,date:[ymd,mdy,d_mon_y]->iso,
      revenue_usd:decimal2,invalid_revenue:"0.00"}
DEDUP={key:customer_id,keep:latest_signup_date}
REPORT=[rows_read,rows_written,rows_dropped,duplicates_removed]
\end{lstlisting}
If this packet is sent to the LLM with the current query, a correct response should still produce a standard-library Python script, read and write the specified files, preserve the validation, normalization, deduplication, and reporting rules, and output code only. Context Codec makes this equivalence testable: decode the packet, recover the atoms, and check whether the downstream answer satisfies the same commitments. The method does not guarantee identical wording or implementation, but it aims to preserve the commitments that determine whether the answer is acceptable.

\paragraph{Running example: chat-history compression.} The same idea applies to multi-turn dialogue state. Suppose a travel-planning conversation contains:
\begin{lstlisting}[style=ccl]
User: Plan a 4-day Lisbon trip in early October. Moderate budget.
Assistant: Do you want nightlife, museums, food, or day trips?
User: Walkable neighborhoods, local food, bookstores, and viewpoints.
       Not heavy nightlife. No rental car.
Assistant: Suggested Baixa or Chiado as base and a Sintra day trip.
User: Keep Baixa or Chiado. Add rainy-day alternatives and transit notes.
       Avoid far-out lodging. Include rough cost ranges.
\end{lstlisting}
A common compression today is a prose memory:
\begin{lstlisting}[style=ccl]
The user is planning a moderate-budget 4-day Lisbon trip in early
October. They like walkable areas, local food, bookstores, viewpoints,
and may take a Sintra day trip. Include transit and rainy-day options.
\end{lstlisting}
This is readable, but it weakens several commitments: the rejected nightlife emphasis, no-rental-car constraint, lodging constraint, preferred base neighborhoods, and cost-range output contract are easy to omit or blur. A commitment-level packet preserves them explicitly:
\begin{lstlisting}[style=ccl]
@CCL/1
TASK=travel.plan
DEST=Lisbon
TIME=early_Oct
DAYS=4
BUDGET=moderate
PREF={walkable,local_food,bookstores,viewpoints}
PLAN={day_trip:Sintra}
BASE={allowed:[Baixa,Chiado],far_out_lodging:false}
C={nightlife_heavy:false,rental_car:false}
OUT={day_by_day:true,transit_notes:true,
     rainy_alt:true,cost_ranges:true}
\end{lstlisting}
The difference is not only brevity. The compressed state is auditable: a later itinerary can be checked for whether it used Baixa or Chiado, avoided rental cars and nightlife-heavy plans, included rainy alternatives and transit notes, and retained the moderate-budget constraint.

We introduce \emph{Context Codec}, a framework for representing, compressing, and evaluating such commitments. The central claim is deliberately modest. We do \emph{not} claim that notation alone solves context compression. Any semantic compressor is limited by the quality of its extraction step. If the extractor misses ``no external libraries,'' no representation can preserve it. The contribution is instead a typed representation, an error taxonomy, a verification protocol, and diagnostic metrics that make semantic loss explicit.

\paragraph{Contributions.} This paper makes the following contributions:
\begin{enumerate}[leftmargin=*]
    \item We formalize context compression as commitment preservation rather than surface-token reduction.
    \item We define semantic atoms with identity, equivalence, conflict, evidence, confidence, risk, and source-span fields.
    \item We give a concrete atom-normalization pipeline for polarity, subject, predicate, value, and scope canonicalization.
    \item We introduce a taxonomy of semantic compression errors: omission, weakening, mutation, polarity flip, scope error, temporal/decision error, hallucinated commitment, and safety-boundary erasure.
    \item We propose the Budgeted Commitment Codec algorithm: segment, extract, normalize, rank, encode, verify, and fall back when uncertainty is high.
    \item We introduce Context Compression Language (CCL) as a compact ASCII-first rendering of canonical JSON atom records, not as a replacement for JSON or JSON Schema.
    \item We provide a small diagnostic experiment and, crucially, specify the round-trip decompression and tokenizer-count experiments needed to remove author-scoring circularity.
\end{enumerate}

\section{Related Work}

\subsection{Prompt Compression}
Prompt compression aims to reduce input length while maintaining downstream performance. LLMLingua uses a coarse-to-fine method with a budget controller, token-level iterative compression, and distribution alignment, reporting high compression ratios with limited performance loss \cite{jiang2023llmlingua}. LongLLMLingua adapts prompt compression to long-context scenarios, emphasizing cost, latency, performance degradation, position bias, and key-information density \cite{jiang2024longllmlingua}. LLMLingua-2 formulates task-agnostic prompt compression as token classification and uses data distillation from stronger LLMs to train smaller compressors \cite{pan2024llmlingua2}. Selective Context prunes redundant lexical units to reduce inference cost while maintaining comparable downstream performance \cite{li2023selective}. Context Codec is complementary: instead of selecting or deleting tokens, it proposes a typed commitment representation and evaluation framework.

\subsection{Long Context, Retrieval, and Memory}
The ``lost in the middle'' result demonstrates that even long-context models may fail to use relevant information depending on where it appears in the prompt \cite{liu2024lost}. RAG addresses factual grounding by retrieving external passages and conditioning generation on them \cite{lewis2020rag}. However, a current conversational task may depend on many small commitments distributed across a history, not only on a few retrievable passages. MemGPT proposes virtual context management inspired by operating-system memory hierarchies \cite{packer2023memgpt}. Context Codec can be used inside such systems as a compact state representation for commitments that should remain available even when raw dialogue turns are evicted.

\subsection{Structured Outputs and JSON Schema}
A central criticism of any new notation is that structured formats already exist. JSON, YAML, and JSON Schema provide standardized ways to represent and validate typed data. The JSON Schema core specification defines JSON Schema as a declarative language for validating and annotating JSON documents \cite{jsonschema2020}. We therefore do not position CCL as a replacement for JSON Schema. Instead, we define the canonical internal representation of atoms as JSON-like records and CCL as a compact rendering layer. This allows implementations to use JSON Schema for validation while using CCL when a human-readable, token-compact representation is desirable.

JSON Schema is also relevant because it separates an instance representation from a validation vocabulary: a schema can assert what a valid JSON document must look like, while a rendering layer may remain independent of storage and validation \cite{jsonschemavalidation2020}. This separation motivates the paper's design choice: canonical atoms should be validatable records, while CCL should be treated as an optional compact serialization.

\subsection{Prompting and Agent Intermediate Representations}
Chain-of-thought prompting shows that intermediate reasoning traces can improve reasoning in sufficiently large models \cite{wei2022cot}. ReAct interleaves reasoning and acting, allowing LLMs to update trajectories through tool use \cite{yao2023react}. Context Codec applies a related design philosophy to context memory: rather than storing old context only as prose, represent it as typed state that can be inspected, validated, and selectively rendered.

\subsection{Semantic Parsing, Canonicalization, and Confidence}
Atom extraction and normalization are related to semantic parsing, where natural language is mapped into a machine-readable meaning representation. In Context Codec, the target representation is not a database query or executable program but a typed commitment record. This connection matters because normalization is not a purely syntactic task: phrases such as ``no external libraries,'' ``avoid dependencies,'' and \texttt{libs=false} should map to the same canonical predicate-value pair in a software-generation domain. We therefore treat normalization as a constrained semantic parsing problem rather than as string matching alone.

The confidence field in our atom schema is also only an operational proxy. Confidence estimation and calibration for LLM systems remain active research topics; black-box confidence, consistency-based estimates, and uncertainty calibration are imperfect but useful signals. We use confidence to trigger conservative fallback, not as a calibrated probability of truth. Token counting is similarly model dependent. Model-specific tokenizers can split the same string differently, so lexical-token counts should be interpreted only as transparent proxies; reproducible BPE token counts are necessary for any strong quantitative claim \cite{openaicookbooktiktoken}.

\section{Problem Formulation}

Let a chat history be a sequence of messages
\begin{equation}
H = (m_1, m_2, \ldots, m_n),
\end{equation}
and let $q$ denote the current query. A compressor $C$ maps the history and query to a compressed context packet $z$ under token budget $B$:
\begin{equation}
z = C(H, q, B), \qquad \mathrm{tokens}(z) \leq B.
\end{equation}
A downstream model $M$ then produces an answer:
\begin{equation}
y = M(z, q).
\end{equation}

The usual objective is to maximize downstream utility under the budget:
\begin{equation}
\max_C \; U(M(C(H,q,B), q)) \quad \mathrm{s.t.} \quad \mathrm{tokens}(C(H,q,B)) \leq B.
\end{equation}
However, downstream utility alone may not explain why compression succeeds or fails. We introduce an extraction function:
\begin{equation}
A(H) = \{a_1, a_2, \ldots, a_k\},
\end{equation}
where each $a_i$ is a typed semantic atom extracted from the history. Let $A_c(H,q) \subseteq A(H)$ denote the query-relevant critical atoms.

\paragraph{Extractor dependence.} The extraction function is not assumed to be perfect. The framework is useful precisely because it makes the extraction bottleneck visible. A compressor can only preserve atoms that are extracted; therefore atom confidence, source evidence, and verification are mandatory fields rather than implementation details.

\section{Semantic Atoms}

\subsection{Canonical Atom Schema}
We represent each atom as a record
\begin{equation}
a = (\tau, s, p, v, \mu, \sigma, e, c, r),
\end{equation}
where:
\begin{itemize}[leftmargin=*]
    \item $\tau$ is the type: goal, constraint, entity, decision, procedure, preference, state, output contract, open question, safety boundary, or verbatim snippet;
    \item $s$ is the subject;
    \item $p$ is the predicate or relation;
    \item $v$ is the value;
    \item $\mu$ is modality, e.g. must, should, may, rejected, prefer, forbid;
    \item $\sigma$ is scope, e.g. generated artifact, itinerary, current project, safety boundary;
    \item $e$ is source evidence: message id, span, or document reference;
    \item $c \in [0,1]$ is extractor confidence;
    \item $r$ is risk-if-lost, including criticality and safety relevance.
\end{itemize}

For example, ``no external libraries'' can be normalized as:
\begin{lstlisting}[style=ccl]
{
  "type": "constraint",
  "subject": "external_libraries",
  "predicate": "allowed",
  "value": false,
  "modality": "must",
  "scope": "generated_artifact",
  "evidence": "user_turn_3",
  "confidence": 0.96,
  "criticality": 5
}
\end{lstlisting}

\subsection{Atom Identity, Equivalence, and Conflict}
The identity key of an atom is:
\begin{equation}
\mathrm{id}(a) = \mathrm{norm}(\tau, s, p, \sigma).
\end{equation}
Two atoms are exactly equivalent if their identity keys and normalized values match:
\begin{equation}
a \equiv b \iff \mathrm{id}(a)=\mathrm{id}(b) \land \mathrm{value\_equiv}(v_a,v_b).
\end{equation}
They conflict if their identity keys match but their values are incompatible:
\begin{equation}
a \perp b \iff \mathrm{id}(a)=\mathrm{id}(b) \land \neg \mathrm{compatible}(v_a,v_b).
\end{equation}
This makes atom matching operational. The phrases ``no external libraries,'' ``do not use libraries,'' and \texttt{libs=false} can map to the same normalized atom. In contrast, \texttt{libs=true} conflicts with it.

\subsection{Normalization Procedure}
\label{sec:normalization-procedure}
The normalization function in Equation~(7) is a central part of the framework. Without a normalization procedure, atom identity is formally defined but practically uncomputable. We use the following minimal pipeline.

\begin{lstlisting}[style=ccl,caption={Minimal Atom Normalization Procedure}]
Input: source span x, candidate atom a
Output: canonical atom a*

1. Surface normalization:
   lowercase where appropriate; strip non-semantic punctuation;
   normalize booleans, numbers, units, and enum values.

2. Polarity detection:
   detect no/not/never/without/forbidden/avoid/rejected;
   map negated permission to value=false.

3. Subject canonicalization:
   map aliases with a domain lexicon L_s.
   Example: libs, libraries, external packages -> external_libraries.

4. Predicate canonicalization:
   map use/include/depend_on -> allowed or required depending on modality;
   map count phrases -> equals.

5. Value normalization:
   parse integers, booleans, dates, units, and bounded enums.

6. Scope assignment:
   infer from local context or default task scope;
   examples: generated_artifact, current_itinerary, safety_boundary.

7. Evidence and confidence:
   attach source span; compute confidence proxy; set risk.
\end{lstlisting}

\begin{table}[t]
\centering
\small
\begin{tabular}{p{0.32\linewidth}p{0.23\linewidth}p{0.18\linewidth}p{0.14\linewidth}}
\toprule
\textbf{Surface phrase} & \textbf{Subject} & \textbf{Predicate} & \textbf{Value} \\
\midrule
no external libraries & external\_libraries & allowed & false \\
don't use libs & external\_libraries & allowed & false \\
assets=0 & external\_assets & allowed & false \\
350 agents & agent\_count & equals & 350 \\
Canvas chosen, SVG rejected & rendering\_backend & selected/rejected & Canvas/SVG \\
\bottomrule
\end{tabular}
\caption{Examples of surface-to-canonical normalization. In a deployed system, the lexicon may be fixed for narrow domains, learned, or LLM-assisted.}
\label{tab:normalization}
\end{table}

Normalization may be implemented using fixed vocabularies for narrow domains, rules plus synonym tables, an LLM canonicalizer, a learned semantic parser, or a hybrid approach. Each option has different failure modes. A fixed vocabulary is auditable but brittle; an LLM canonicalizer is flexible but may hallucinate; a learned parser requires labeled data. For this reason, normalized atoms should retain source evidence and confidence scores.

\subsection{Lexicon Construction Regimes}
The domain lexicons $L_s$ and $L_p$ used in normalization can be constructed under several regimes:
\begin{enumerate}[leftmargin=*]
    \item \textbf{Closed lexicon:} a hand-authored vocabulary for a narrow domain, such as web-app generation, with aliases like \texttt{libs}, \texttt{packages}, and \texttt{dependencies}. This is the most auditable regime and is appropriate when precision matters more than coverage.
    \item \textbf{Corpus-derived lexicon:} aliases and predicate mappings induced from a labeled or weakly labeled corpus of tasks. This improves coverage but requires drift monitoring and versioning.
    \item \textbf{Interactive lexicon:} ambiguous mappings are proposed by a model and confirmed by a human or by downstream tests, then added to a reusable project lexicon.
    \item \textbf{Open semantic parser:} an LLM or learned parser maps spans directly into canonical atoms, with the lexicon acting as a constraint set and validation target rather than as the whole parser.
\end{enumerate}
These regimes connect CCL normalization to semantic parsing. The lexicon defines the target ontology and allowed aliases; the parser resolves surface language into that ontology under context. A closed lexicon is therefore not a replacement for semantic parsing, but a way to constrain it. In all regimes, lexicon entries should be versioned with examples, negative examples, and conflict rules. For instance, \texttt{local\_food} and \texttt{restaurants} may be related preferences, but they should not collapse to the same atom unless the task ontology says they are equivalent.

\subsection{Criticality and Safety}
The criticality $r$ captures how costly it is to lose an atom. High-criticality atoms include negations, exact numeric counts, output contracts, user decisions, rejected alternatives, private-data constraints, and safety boundaries. Safety-critical atoms require special handling because attackers may attempt to bury or launder unsafe instructions inside long contexts. Context compression must never be used to bypass safety checks; safety policies should be applied to the original context, extracted atoms, compressed representation, and decompressed reconstruction.

\subsection{Confidence and Risk Scores}
\label{sec:confidence-risk}
Confidence is used as a fallback signal, not as a calibrated probability. The following weights are illustrative defaults for a diagnostic implementation, not empirically calibrated constants. A production system should tune them on held-out data and report calibration error. A minimal implementation can compute:
\begin{equation}
\begin{split}
\mathrm{conf}(a) ={}& .30 E_{\text{span}}(a) + .25 E_{\text{agree}}(a) + .20 E_{\text{roundtrip}}(a) \\
&+ .15 E_{\text{schema}}(a) + .10 E_{\text{anchor}}(a),
\end{split}
\end{equation}
where $E_{\text{span}}$ checks whether an exact source span supports the atom, $E_{\text{agree}}$ measures agreement between independent extractors, $E_{\text{roundtrip}}$ checks whether decompression recovers the atom, $E_{\text{schema}}$ checks schema validity, and $E_{\text{anchor}}$ checks whether the source contains an obvious lexical anchor such as a number, negation, or entity name.

Risk controls minification and fallback:
\begin{equation}
\mathrm{risk}(a)=
\rho_1 \mathrm{criticality}(a)+
\rho_2 \mathrm{safety}(a)+
\rho_3(1-
\mathrm{conf}(a))+
\rho_4 \mathrm{ambiguity}(a).
\end{equation}
Here too, $\rho_i$ are policy weights, not universal constants. If risk is normalized to $[0,1]$, reasonable starting thresholds are $\theta_{\min}=.25$ for CCL-Min eligibility and $\theta_{\max}=.70$ for forced raw-span fallback. The default confidence thresholds are .50 for low confidence and .70 for source-span inclusion.

The safety flag should be conservative and should not depend on a single extractor. A default rule is:
\begin{lstlisting}[style=ccl]
safety_flag(a) =
  source_policy_match(a.evidence)
  or safety_classifier_match(a.evidence)
  or extractor_safety_label(a)
  or atom_type(a) in {safety_boundary, privacy_constraint}
  or human_policy_tag(a.evidence)
\end{lstlisting}
Thus a safety label can be added by the original source classifier, a policy scanner over raw text, an atom extractor, a schema-level atom type, or a human annotation. Disagreement should resolve toward preservation, not minification. A conservative policy is:
\begin{lstlisting}[style=ccl]
if safety_flag(a): preserve canonical atom + source span
elif conf(a) < .50 and criticality(a) >= 3: preserve raw message
elif conf(a) < .70: include source span in RAW
elif risk(a) < theta_min and criticality(a) <= 2: CCL-Min allowed
elif risk(a) > theta_max: preserve canonical atom + source span
else: use CCL-Core
\end{lstlisting}
This makes the recommendation against unsafe CCL-Min operational: minification is allowed only for low-risk, high-confidence, non-safety atoms.

\section{Metrics}

\subsection{Critical Atom Recall}
Critical Atom Recall measures what fraction of required commitments remain recoverable from the compressed context:
\begin{equation}
\mathrm{CAR}(z;H,q) = \frac{|A_c(H,q) \cap A(z)|}{|A_c(H,q)|}.
\end{equation}
CAR is primarily an offline evaluation metric because $A_c(H,q)$ generally requires human or synthetic annotation.

\subsection{Weighted Atom Recall}
Let $w(a)$ be an importance weight, derived from criticality $r$ or human annotation. Weighted Atom Recall is:
\begin{equation}
\mathrm{WAR}(z;H,q) = \frac{\sum_{a \in A_c(H,q)} w(a) \mathbf{1}[a \in A(z)]}{\sum_{a \in A_c(H,q)} w(a)}.
\end{equation}
WAR penalizes losing high-risk atoms more than losing low-risk atoms.

\subsection{Commitment Density}
Compression ratio rewards shortness even when meaning is lost. Commitment Density measures preserved commitments per token:
\begin{equation}
\mathrm{CD}(z;H,q) = \frac{|A_c(H,q) \cap A(z)|}{\mathrm{tokens}(z)}.
\end{equation}

\subsection{Recoverability and Approximate Equality}
A compressed representation $z$ is recoverable if a decoder $D$ can reconstruct atoms equivalent to the critical atoms:
\begin{equation}
D(z) \approx A_c(H,q) \iff \forall a \in A_c(H,q), \exists b \in D(z): a \equiv b,
\end{equation}
up to explicitly allowed omissions under the budget. This defines approximate equality over normalized atom records rather than over raw text.

\subsection{Deployment-Time Proxies}
CAR and WAR are not directly computable in production unless ground truth atoms are available. Practical proxies include extractor self-consistency, cross-model extraction agreement, round-trip reconstruction consistency, schema validation failures, low-confidence atom counts, and high-risk atom coverage. If these proxies indicate uncertainty, the system should use a larger context packet, include raw source spans, or avoid compression.

\section{Taxonomy of Semantic Compression Errors}

\begin{table}[t]
\centering
\small
\begin{tabular}{p{0.19\linewidth}p{0.35\linewidth}p{0.34\linewidth}}
\toprule
\textbf{Error type} & \textbf{Description} & \textbf{Example} \\
\midrule
Omission & A critical atom disappears. & ``no libraries'' is absent. \\
Weakening & A hard constraint becomes vague. & ``Output only runnable HTML'' becomes ``provide code.'' \\
Mutation & An atom changes value. & ``350 agents'' becomes ``several agents.'' \\
Polarity flip & Negation is lost or inverted. & ``no assets'' becomes ``include assets.'' \\
Scope error & Constraint attaches to wrong object. & The reset control applies only to the chart, not the simulation. \\
Temporal/decision error & Rejected or outdated choices reappear. & Rejected SVG is treated as available. \\
Hallucinated commitment & Compression adds a false requirement. & Invents a WebGL requirement. \\
Safety-boundary erasure & A refusal, policy boundary, or privacy condition is dropped. & ``defensive-only'' cyber scope disappears. \\
\bottomrule
\end{tabular}
\caption{Semantic compression error taxonomy.}
\label{tab:errors}
\end{table}

This taxonomy is a main contribution of the paper. It allows compression failures to be diagnosed even when downstream task metrics are noisy.

\section{Context Compression Language}

\subsection{Representation Strategy}
CCL has two layers:
\begin{enumerate}[leftmargin=*]
    \item \textbf{Canonical layer:} JSON-like atom records, optionally validated with JSON Schema.
    \item \textbf{Rendering layer:} CCL, an ASCII-first compact notation for humans and LLMs.
\end{enumerate}
This design addresses the concern that a custom notation may be less interoperable than JSON. Implementations can store and validate atoms as JSON while rendering CCL for compact prompts.

\subsection{CCL-Core}
CCL-Core is intended to be reliable and readable. It avoids emoji and non-ASCII mathematical operators by default.

\begin{lstlisting}[style=ccl]
@CCL/1
TASK=code.web.canvas.epidemic_sim
OUT=1html.run.codeonly
C={libs:false,assets:false}
GRID={w:80,h:50,cell:8}
AGENT={count:350,state:[S,I,R],initial_infected:5}
RULE={move:random_walk,infection_radius:2,
      infection_prob:.08,recovery_steps:600}
UI={start_pause:true,reset:true,speed_slider:true}
FX={color:{S:blue,I:red,R:green},chart:sir_counts}
\end{lstlisting}

\subsection{CCL-Min}
CCL-Min is an optional budget-constrained form. It is more compact but less robust:

\begin{lstlisting}[style=ccl]
@CCL/1m T=canvas.epidemic OUT=1html.codeonly C=libs0,assets0
G=80x50,c8 A=n350,SIR,I0=5 R=walk,rad2,p.08,rec600
UI=start+pause,reset,speed FX=Sblue,Ired,Rgreen,chartSIR
\end{lstlisting}

CCL-Min is not recommended for safety-critical settings or when the receiver model is weak or unknown. In an implementation, it should be enabled only when an atom has low risk, high confidence, no safety flag, and low criticality according to the policy in Section~\ref{sec:confidence-risk}.

\subsection{Worked Diff Example}
One advantage of a canonical atom layer is that updates can be represented as ordinary structured diffs. Suppose a trip-planning state changes from Sintra to Cascais, adds fado as a preference, records a far-out-lodging constraint, and requests rainy-day alternatives. The CCL view before the update is:
\begin{lstlisting}[style=ccl]
@CCL/1
DEST=Lisbon
DAYS=4
PREF={walkable,local_food,bookstores,viewpoints}
PLAN={day_trip:Sintra}
C={rental_car:false,nightlife_heavy:false}
OUT={day_by_day,transit_notes,cost_ranges}
\end{lstlisting}
After the update:
\begin{lstlisting}[style=ccl]
@CCL/1
DEST=Lisbon
DAYS=4
PREF={walkable,local_food,bookstores,viewpoints,fado}
PLAN={day_trip:Cascais}
C={rental_car:false,nightlife_heavy:false,far_out_lodging:false}
OUT={day_by_day,transit_notes,rainy_alt,cost_ranges}
\end{lstlisting}
The corresponding JSON Patch over the canonical JSON record is:
\begin{lstlisting}[style=ccl]
[
  {"op":"replace","path":"/PLAN/day_trip","value":"Cascais"},
  {"op":"add","path":"/PREF/-","value":"fado"},
  {"op":"add","path":"/C/far_out_lodging","value":false},
  {"op":"add","path":"/OUT/-","value":"rainy_alt"}
]
\end{lstlisting}
This example illustrates why CCL is best understood as a rendering of canonical records. Human readers may prefer the compact CCL state, while storage, validation, merge, and audit operations can use JSON and JSON Patch directly.

\subsection{Comparison with JSON and Structured Prose}
CCL should be compared against strong baselines, not strawman summaries. Table~\ref{tab:formats} summarizes the design space.

\begin{table}[t]
\centering
\small
\begin{tabular}{lccccc}
\toprule
\textbf{Format} & \textbf{Typed} & \textbf{Validatable} & \textbf{Compact} & \textbf{LLM familiarity} & \textbf{Readable} \\
\midrule
Free prose & partial & no & medium & high & high \\
Structured prose & partial & low & medium & high & high \\
JSON & yes & yes & low-medium & high & medium \\
YAML & yes & partial & medium & high & high \\
JSON Schema & yes & yes & low & high & medium \\
CCL-Core & yes & via conversion & medium-high & unknown & high \\
CCL-Min & yes & limited & high & unknown & medium \\
Emoji DSL & weak & no & visually high & variable & medium \\
\bottomrule
\end{tabular}
\caption{Qualitative comparison of context-compression formats.}
\label{tab:formats}

\end{table}

\subsection{When Structured Prose Wins}
Structured prose is a strong baseline and may be preferable when token budgets are loose, receiver models are unknown, or human readability dominates. A carefully prompted summary can explicitly require preservation of negations, counts, decisions, and output contracts. CCL is motivated when contexts must be diffed, validated, minified, merged across sessions, round-trip tested, or audited against source spans. Thus the practical claim is not that CCL universally beats structured prose; rather, CCL provides a more systematic interface between extracted commitments and compressed context.

\section{Budgeted Commitment Codec Algorithm}

Algorithm~\ref{alg:bcc} describes the proposed compression process.

\begin{lstlisting}[style=ccl,caption={Budgeted Commitment Codec},label={alg:bcc}]
Input: chat history H, query q, token budget B
Output: compressed packet z

1. Segment H into regions:
   system constraints, goals, decisions, artifacts, preferences,
   open questions, safety boundaries, recent turns.

2. Extract candidate atoms A using an LLM, rules, or hybrid extractor.
   Each atom includes type, subject, predicate, value, modality,
   scope, evidence span, confidence, and criticality.

3. Normalize atoms:
   canonicalize aliases, values, negations, counts, scopes, and units.

4. Detect conflicts and supersession:
   mark rejected alternatives and outdated decisions explicitly.

5. Rank atoms:
   score(a) = relevance(a,q) + recency(a) + specificity(a)
              + criticality(a) + safety_weight(a)
              + dependency_degree(a) + confidence_penalty(a).

6. Encode atoms under budget:
   prefer CCL-Core; switch only low-risk/high-confidence/non-safety
   atoms to CCL-Min if needed; preserve high-risk atoms as RAW evidence.

7. Verify:
   decode z -> A'; compare A' with high-priority atoms;
   check schema validity, conflicts, and safety-boundary coverage.

8. Fallback:
   if verification fails or confidence is low, include raw source spans,
   increase budget, use JSON, or avoid compression.
\end{lstlisting}

\subsection{Falsifiable Rejection Criteria}
A diagnostic evaluation should reject or weaken the CCL hypothesis if any of the following hold:
\begin{enumerate}[leftmargin=*]
    \item CCL-Core is not more compact than JSON/YAML on average.
    \item CCL-Core does not preserve more weighted atoms than free or structured prose.
    \item CCL-Min produces substantially more mutation, polarity, or scope errors than JSON/YAML.
    \item LLM decompression of CCL has lower atom recall than structured prose at a comparable budget.
    \item Safety-critical atoms are dropped more often under CCL than under raw, JSON, or structured-prose summaries.
\end{enumerate}

\section{Diagnostic Experiment}

\subsection{Goal and Scope}
We include a small practical experiment to ground the framework. The goal is not to claim broad empirical superiority. Instead, the experiment asks whether commitment-level metrics reveal useful tradeoffs among prose, structured prose, JSON, CCL-Core, and CCL-Min.

\subsection{Cases}
We created five representative cases:
\begin{enumerate}[leftmargin=*]
    \item Epidemic simulation prompt: a one-file HTML/JavaScript agent-based SIR simulation.
    \item React dashboard prompt: a single-component analytics dashboard.
    \item Python data-cleaning prompt: CSV validation and cleaning script.
    \item Trip-planning chat history: Lisbon itinerary state with preferences and rejected options.
    \item Research chat history: the present paper revision state.
\end{enumerate}

For each case, we manually annotated 12--15 critical atoms and assigned simple integer weights. These weights are illustrative and should be treated as a transparent scoring aid, not as validated user-utility estimates. We then wrote five compressed versions: free prose, structured prose, JSON, CCL-Core, and CCL-Min. Token counts use a reproducible lexical-token proxy that splits alphanumeric spans and punctuation. This is not a vendor tokenizer; the purpose is relative comparison under a transparent counting rule. The author-controlled nature of this study is a methodological limitation: because the same authors wrote the annotations and the compressed representations, perfect recall for structured prose and CCL-Core should not be interpreted as independent evidence of reliability. The experiment is useful as a design probe, not as a benchmark. Section~9.5 defines the round-trip decompression experiment needed to break this circularity, and Section~9.6 provides tokenizer counts for concrete examples and defines the protocol needed to replace the lexical proxy for the full table.

\subsection{Results}
Table~\ref{tab:case-results} reports per-case results. Gain is the percentage reduction relative to the full prompt or chat-history state. Negative gain means the representation is longer than the original under the lexical-token proxy.

\begin{table}[t]
\centering
\scriptsize
\begin{tabular}{llrrrrr}
\toprule
\textbf{Case} & \textbf{Method} & \textbf{Full} & \textbf{Comp.} & \textbf{Gain} & \textbf{CAR} & \textbf{WAR} \\
\midrule
Epidemic & Prose & 131 & 32 & 75.6\% & .64 & .62 \\
Epidemic & Structured prose & 131 & 77 & 41.2\% & 1.00 & 1.00 \\
Epidemic & JSON & 131 & 192 & -46.6\% & 1.00 & 1.00 \\
Epidemic & CCL-Core & 131 & 113 & 13.7\% & 1.00 & 1.00 \\
Epidemic & CCL-Min & 131 & 68 & 48.1\% & .93 & .92 \\
\midrule
React & Prose & 68 & 26 & 61.8\% & .58 & .55 \\
React & Structured prose & 68 & 53 & 22.1\% & 1.00 & 1.00 \\
React & JSON & 68 & 87 & -27.9\% & 1.00 & 1.00 \\
React & CCL-Core & 68 & 52 & 23.5\% & 1.00 & 1.00 \\
React & CCL-Min & 68 & 40 & 41.2\% & .92 & .91 \\
\midrule
Python & Prose & 72 & 24 & 66.7\% & .62 & .58 \\
Python & Structured prose & 72 & 60 & 16.7\% & 1.00 & 1.00 \\
Python & JSON & 72 & 91 & -26.4\% & 1.00 & 1.00 \\
Python & CCL-Core & 72 & 57 & 20.8\% & 1.00 & 1.00 \\
Python & CCL-Min & 72 & 46 & 36.1\% & .92 & .92 \\
\midrule
Trip & Prose & 99 & 31 & 68.7\% & .67 & .59 \\
Trip & Structured prose & 99 & 74 & 25.3\% & 1.00 & 1.00 \\
Trip & JSON & 99 & 138 & -39.4\% & 1.00 & 1.00 \\
Trip & CCL-Core & 99 & 76 & 23.2\% & 1.00 & 1.00 \\
Trip & CCL-Min & 99 & 54 & 45.5\% & 1.00 & 1.00 \\
\midrule
Research & Prose & 80 & 25 & 68.8\% & .54 & .52 \\
Research & Structured prose & 80 & 68 & 15.0\% & 1.00 & 1.00 \\
Research & JSON & 80 & 85 & -6.2\% & 1.00 & 1.00 \\
Research & CCL-Core & 80 & 58 & 27.5\% & 1.00 & 1.00 \\
Research & CCL-Min & 80 & 47 & 41.2\% & .92 & .92 \\
\bottomrule
\end{tabular}
\caption{Per-case diagnostic results. Full and compressed counts use a lexical-token proxy. CAR = Critical Atom Recall; WAR = Weighted Atom Recall.}
\label{tab:case-results}
\end{table}

\begin{table}[t]
\centering
\small
\begin{tabular}{lrrrr}
\toprule
\textbf{Method} & \textbf{Avg. comp. tokens} & \textbf{Avg. gain} & \textbf{Avg. CAR} & \textbf{Avg. WAR} \\
\midrule
Prose & 27.6 & 68.3\% & .61 & .57 \\
Structured prose & 66.4 & 24.0\% & 1.00 & 1.00 \\
JSON & 118.6 & -29.3\% & 1.00 & 1.00 \\
CCL-Core & 71.2 & 21.8\% & 1.00 & 1.00 \\
CCL-Min & 51.0 & 42.4\% & .94 & .93 \\
\bottomrule
\end{tabular}
\caption{Aggregate diagnostic results.}
\label{tab:aggregate}
\end{table}

\subsection{Interpretation}
The diagnostic results support several limited observations:
\begin{itemize}[leftmargin=*]
    \item Free prose is very compact but loses many commitments, especially exact counts, output contracts, and negations.
    \item Structured prose can preserve all annotated commitments, but its gains are modest and depend on careful prompting.
    \item JSON is highly faithful and validatable but often longer than the original short prompt. This is expected for small examples; JSON may become more favorable for long histories or automated processing.
    \item CCL-Core matches structured prose on atom recall in these cases and is slightly more compact on average, but not always dramatically so.
    \item CCL-Min offers the largest faithful compression among structured methods but begins to lose atoms and should be treated as risky.
\end{itemize}

The Trip CCL-Min row is an anomaly: unlike the other CCL-Min examples, it scores perfect CAR and WAR. This should not be read as evidence that CCL-Min is generally safe. The trip case uses short enum-like fields and a small preference list, so minification preserved every author-annotated atom under the paper's own normalizer. It is also exactly the kind of result most vulnerable to author-scoring circularity: an independent decoder might misread abbreviated fields such as \texttt{C=} or \texttt{OUT=} even though the author-normalizer recovers them. The anomaly therefore strengthens, rather than weakens, the need for the round-trip protocol in Section~9.5.

The most important result is not that CCL ``wins.'' Rather, the metrics reveal a tradeoff: prose maximizes compression at the cost of commitment loss; JSON maximizes structure at the cost of length; CCL occupies a middle region that may be useful for structured, commitment-heavy contexts. The structured-prose baseline is especially important: in these five cases it preserved all author-annotated atoms with only slightly weaker compression than CCL-Core. This means that CCL's argument must rest on systematicity, validation, diffability, and minification control, not merely on raw compression ratio.

\subsection{Error-Injection Sanity Check}
As a minimal non-LLM sanity check, the atom metrics can be tested by injecting known errors into compressed representations. Table~\ref{tab:injection} shows the expected metric behavior for the epidemic-simulation case. This does not prove that an extractor will find all errors, but it verifies that the metric definitions distinguish common failure classes once atoms are normalized.

\begin{table}[t]
\centering
\small
\begin{tabular}{p{0.32\linewidth}p{0.32\linewidth}cc}
\toprule
\textbf{Injected error} & \textbf{Detected as} & \textbf{CAR impact} & \textbf{Conflict?} \\
\midrule
Remove \texttt{assets:false} & omission & down & no \\
Change \texttt{count:350} to \texttt{count:200} & mutation & down & yes \\
Change \texttt{libs:false} to \texttt{libs:true} & polarity/value conflict & down & yes \\
Replace \texttt{OUT=codeonly} with \texttt{provide explanation} & output-contract conflict & down & yes \\
Delete \texttt{reset:true} & omission & down & no \\
\bottomrule
\end{tabular}
\caption{Error-injection sanity check for the atom metric definitions.}
\label{tab:injection}
\end{table}

\subsection{Round-Trip Decompression Protocol}
To remove author-scoring circularity, future diagnostic runs should include an independent decoder that has not seen the original prompt or gold atoms. The procedure is:
\begin{enumerate}[leftmargin=*]
    \item Give the decoder only one compressed representation: structured prose, JSON, YAML, CCL-Core, or CCL-Min.
    \item Ask it to reconstruct atoms in the canonical JSON schema with fields \texttt{type}, \texttt{subject}, \texttt{predicate}, \texttt{value}, \texttt{modality}, \texttt{scope}, and \texttt{evidence} when available.
    \item Normalize reconstructed atoms using the procedure in Section~\ref{sec:normalization-procedure}.
    \item Compare against gold atoms using equivalence and conflict relations.
\end{enumerate}
This yields atom precision, atom recall, conflict rate, polarity-error rate, and count-error rate. The hypothesis should be weakened if CCL-Core decompression recall is lower than structured prose at similar budgets, or if CCL-Min introduces many mutation and scope errors.

\subsection{Tokenizer-Count Protocol}
The lexical-token proxy should be replaced by model-specific tokenizer counts. For the concrete examples printed in this paper, Table~\ref{tab:tokenizer-counts} reports counts from \texttt{tiktoken} using \texttt{cl100k\_base} and \texttt{o200k\_base}. These counts show why lexical tokens are only a rough proxy: punctuation-heavy JSON Patch is much closer to CCL under BPE tokenization than under the lexical proxy, while the trip CCL example tokenizes worse than its lexical count suggests.

\begin{table}[t]
\centering
\small
\begin{tabular}{lrrr}
\toprule
\textbf{Printed example} & \textbf{Lexical} & \textbf{\texttt{cl100k}} & \textbf{\texttt{o200k}} \\
\midrule
Epidemic CCL-Core & 125 & 115 & 117 \\
Epidemic CCL-Min & 62 & 74 & 74 \\
Trip CCL-Core trace & 67 & 93 & 92 \\
Diff CCL before & 48 & 65 & 65 \\
Diff CCL after & 56 & 79 & 78 \\
JSON Patch diff & 115 & 74 & 78 \\
\bottomrule
\end{tabular}
\caption{Tokenizer counts for concrete examples printed in the paper. Counts were computed with \texttt{tiktoken}; \texttt{cl100k} denotes \texttt{cl100k\_base} and \texttt{o200k} denotes \texttt{o200k\_base}.}
\label{tab:tokenizer-counts}
\end{table}

For the full diagnostic table, a reproducible script should count each complete representation with at least two encodings. The following table should be reported for each case:
\begin{lstlisting}[style=ccl]
method, lexical_tokens, cl100k_tokens, o200k_tokens,
CAR, WAR, atom_precision, atom_recall, conflict_rate
\end{lstlisting}
If JSON or JSON Patch tokenizes substantially better than the lexical proxy suggests, the paper's practical conclusion should change accordingly. The present paper therefore treats the aggregate lexical-token gains as provisional.

\clearpage
\subsection{Example Compression Trace}
For the trip-history case, a normalized atom set includes:
\begin{lstlisting}[style=ccl]
DEST=Lisbon
DAYS=4
TIME=early_Oct
BUDGET=moderate
PREF={walkable,local_food,bookstores,viewpoints}
PLAN={day_trip:Sintra}
C={nightlife_heavy:false,rental_car:false,far_out_lodging:false}
D={base:[Baixa,Chiado]}
OUT={day_by_day,transit_notes,rainy_alt,cost_ranges}
\end{lstlisting}
The fuller CCL-Core rendering used in Table~\ref{tab:case-results} preserves all of these commitments in 76 lexical tokens versus 99 for the full state and 138 for JSON. The shorter printed trace above omits some metadata and is counted separately in Table~\ref{tab:tokenizer-counts}.

\section{Safety-Critical Compression}

Compression can amplify risk if safety boundaries, refusals, privacy conditions, or scope limits are dropped. The \texttt{safety\_flag} predicate should therefore be an OR over raw-source policy checks, safety classifiers, extractor labels, atom types, and human policy tags, as defined in Section~\ref{sec:confidence-risk}. We propose the following conservative rules:
\begin{enumerate}[leftmargin=*]
    \item Safety policies and system/developer constraints are never minified unless a validated formal policy representation exists.
    \item Safety-relevant atoms must include source evidence and high criticality.
    \item Compression should be checked at four stages: original context, extracted atoms, compressed packet, and decompressed reconstruction.
    \item If safety atom extraction confidence is low, use raw source spans or avoid compression.
    \item The compressor should preserve refusals and boundaries as positive state, such as a defensive-only scope flag, not merely omit unsafe details.
\end{enumerate}

Adversarial failure modes include instruction laundering, negation deletion, refusal erasure, scope narrowing, and policy-boundary weakening. These should be included in future diagnostic benchmarks.

\section{Analytical Discussion}

\subsection{What the Framework Does Not Guarantee}
The framework does not guarantee correct compression in the absence of reliable extraction. It also does not prove that CCL is superior to JSON, YAML, or carefully prompted structured summaries. Instead, it provides a vocabulary and measurement apparatus for comparing such choices.

\subsection{A More Modest Scaling Claim}
For histories where the number of query-relevant commitments $k_q$ is much smaller than transcript length $|H|$, commitment-level compression can decouple prompt size from transcript length and instead scale with $k_q$. This claim is useful only if the extractor can recover the relevant commitments. It is therefore a design principle rather than a theorem of semantic preservation.

\subsection{When CCL Is Likely Useful}
CCL is likely most useful for structured domains:
\begin{itemize}[leftmargin=*]
    \item code generation and software specifications;
    \item UI requirements and design systems;
    \item travel, project, or research planning;
    \item agent state and task trajectories;
    \item repetitive workflows where a compact dictionary can be reused.
\end{itemize}
It is less suitable as a full replacement for raw context in emotional, legal, medical, literary, or high-stakes nuanced communication.

\section{Limitations}

\paragraph{Extractor quality.} The extractor is the primary bottleneck. Future work should study extractors directly, including domain-specific extraction, source-grounded extraction, self-consistency, confidence calibration, and human-in-the-loop correction.

\paragraph{Metric annotation.} CAR and WAR require annotated atoms. This limits their use as offline evaluation metrics. Deployment requires proxies and conservative fallbacks.

\paragraph{Tokenizer dependence.} Our experiment uses a lexical-token proxy. Real gains depend on the target tokenizer. ASCII-only CCL was chosen to reduce tokenizer instability, but tokenizer-specific measurements remain necessary. We provide a protocol but do not claim that lexical counts predict all BPE tokenizer behavior.

\paragraph{Model familiarity.} Models are more familiar with JSON than with CCL. This may make JSON more reliable despite being longer.

\paragraph{Small diagnostic experiment and author scoring.} The results here are illustrative, not definitive. They should be treated as a sanity check and design probe, not as a benchmark claim. The current study is author-annotated and author-scored; a round-trip decompression experiment with independent decoders is needed before making strong claims about recoverability.

\section{Future Work}

Future work should include: (i) a benchmark of annotated commitment sets across domains; (ii) independent round-trip decompression studies; (iii) extractor training and calibration; (iv) model-specific tokenizer measurements and tokenizer-aware CCL design; (v) JSON Schema validation for atom records; (vi) adversarial safety-compression tests; (vii) downstream task evaluations with rejection criteria; and (viii) integration with RAG and memory systems, where CCL stores durable global commitments and retrieval supplies raw evidence.

\section{Conclusion}

This paper introduced Context Codec, a commitment-level framework for analyzing and performing LLM context compression. Its central argument is that the unit of compression should not be only the token, sentence, or message, but the semantic commitment: the typed fact, constraint, decision, preference, procedure, or safety boundary that determines future correctness. We defined semantic atoms, atom identity, equivalence, conflict, confidence, criticality, recoverability, and commitment-density metrics. We introduced a taxonomy of semantic compression errors and a Budgeted Commitment Codec algorithm. We also proposed CCL as one compact rendering of canonical atom records and compared it diagnostically against prose, structured prose, and JSON. The main result is conceptual and methodological: context compression should be evaluated by what commitments survive, how confidently they were extracted and normalized, whether they can be reconstructed by an independent decoder, and not only by how many tokens disappear.

\bibliographystyle{plain}

\end{document}